\documentclass[a4paper]{article}

\usepackage{amsmath,amsfonts,amssymb}
\usepackage{graphicx}
\usepackage{booktabs}
\usepackage{subfig}
\usepackage[colorlinks=true, allcolors=blue]{hyperref}
\usepackage{enumitem}

\title{An Intrinsically Explainable Approach to Detecting Vertebral Compression Fractures in CT Scans via Neurosymbolic Modeling}

\author{
  Blanca Inigo\textsuperscript{1}, 
  Yiqing Shen\textsuperscript{1}, 
  Benjamin D. Killeen\textsuperscript{1}, \\
  Michelle Song\textsuperscript{2}, 
  Axel Krieger\textsuperscript{2}, 
  Christopher Bradley\textsuperscript{3}, \\
  Mathias Unberath\textsuperscript{1} \\
  \\
  \textsuperscript{1}Department of Computer Science, Johns Hopkins University, Baltimore, MD \\
  \textsuperscript{2}Department of Mechanical Engineering, Johns Hopkins University, Baltimore, MD \\
  \textsuperscript{3}Department of Interventional Radiology, Johns Hopkins University, Baltimore, MD
}

\newlist{inlist}{enumerate*}{1}
\setlist[inlist]{label=(\arabic*)}
\newlist{inlistalpha}{enumerate*}{1}
\setlist[inlistalpha]{label=(\alph*)}
\newcommand{\keywords}[1]{\textbf{Keywords:} #1}

\usepackage[backend=biber,maxnames=1,minnames=1,style=numeric,sorting=none]{biblatex}
\begin{document} 
\maketitle

\begin{abstract}

Vertebral compression fractures (VCFs) are a common and potentially serious consequence of osteoporosis. Yet, they often remain undiagnosed.
Opportunistic screening, which involves automated analysis of medical imaging data acquired primarily for other purposes, is a cost-effective method to identify undiagnosed VCFs.
In high-stakes scenarios like opportunistic medical diagnosis, model interpretability is a key factor for the adoption of AI recommendations. Rule-based methods are inherently explainable and closely align with clinical guidelines, but they are not immediately applicable to high-dimensional data such as CT scans. 
To address this gap, we introduce a neurosymbolic approach for VCF detection in CT volumes.
The proposed model combines deep learning (DL) for vertebral segmentation with a shape-based algorithm (SBA) that analyzes vertebral height distributions in salient anatomical regions. 
This allows for the definition of a rule set over the height distributions to detect VCFs. 
Evaluation of VerSe19 dataset shows that our method achieves an accuracy of 96\% and a sensitivity of 91\% in VCF detection.
In comparison, a black box model, DenseNet, achieved an accuracy of 95\% and sensitivity of 91\% in the same dataset. 
Our results demonstrate that our intrinsically explainable approach can match or surpass the performance of black box deep neural networks while providing additional insights into why a prediction was made.
This transparency can enhance clinician's trust 
 thus, supporting more informed decision-making in VCF diagnosis and treatment planning.

\end{abstract}


\section{INTRODUCTION}
\label{sec:intro}  %

Vertebral compression fractures (VCFs) are a significant health concern, particularly in aging populations.\cite{simon2003prevention, alexandru2012evaluation, lee2022clinical} 
As the most common complication of osteoporosis,\cite{alsoof2022diagnosis} VCFs affect more than 700,000 Americans annually\cite{barr2000percutaneous, mccarthy2016diagnosis, alexandru2012evaluation, wong2013vertebral} and have a global incidence rate of 10.7 per 1000 women and 5.7 per 1000 men as of 2012.\cite{alexandru2012evaluation}. %
Automated detection of VCFs in computed tomography (CT) images is an important step toward low-cost opportunistic screening of VCFs. %
It enables straightforward screening of CT images acquired in the course of routine care, \cite{Engelke23} and supports large-scale extraction of the desired cohort from existing image databases. Ultimately, these cohorts can be used to generate large-scale \emph{in silico} simulations of VCF cases in other medical image modalities \cite{killeen2023silico, yang2020mri}, facilitating further research and development in the field.

In recent years, deep neural networks (DNNs) have yielded significant improvements to VCF detection and classification,\cite{bar2017compression, bendtsen2024opportunistic} but the lack of explainability in these algorithms is regarded as a barrier to their real-world implementation and adoption.\cite{pickhardt2023opportunistic} One avenue for explainability is the use of saliency maps to identify image regions that most influenced a DNN's output.\cite{borys2023explainable} However, recent work suggests that saliency-based explainability is vulnerable to imperceptible perturbations in the input, which cause a DNN to reverse its decision without affecting the corresponding saliency map.\cite{zhang2024revisiting} %
Other promising strategies leverage deep learning techniques with interpretable algorithms, ensuring the decision-making process adheres to standardized medical guidelines. \cite{haomin2024trauma, Haomin2023spleen, Ana2021Trauma} There is, however, an unmet need for more interpretable VCF diagnosis algorithms based on CT imaging.

Neurosymbolic AI, which combines the strengths of neural networks and symbolic AI, has been explored to create more interpretable AI systems\cite{Haomin2021Cito}. There is a growing interest in evaluating user trust and adherence given different explainability mechanisms\cite{Cata2024Gaucoma, Cata2024Strep}. In the scope of VCF detection, Burns et al.\cite{RSNA2017VCF} and Baum et al.\cite{Baum2014VCF} automated the extraction of vertebral height parameters and used them to detect fractures, employing a machine learning model and statistical tests, respectively. However, these approaches still fall short of providing a fully transparent and interpretable solution.
To address this need, we propose an intrinsically explainable model that extracts symbolic representations of knowledge and defines logical rules to accurately detect VCFs. Our approach generates standardized 2D height maps for each vertebra and computes statistical measurements from multiple sections of the map. Unlike traditional methods that focus on anterior, middle, and posterior heights (AH, MH, PH)\cite{Lenchik2004AJR, Baum2014VCF, RSNA2017VCF, verseVCFDet}, our model captures height measurements across the entire axial plane of the vertebra. This allows for a more complete representation of vertebral structures and deformations.
Using these parameters, a predefined 2-rule set indicates whether the vertebra is moderate or severely fractured, ensuring generalization and interpretability. 

\section{METHODS}
\label{sec:methods} 

\subsection{Data processing and parameter extraction overview}
\label{sec:overview}

This method aims to provide an interpretable decision-making pipeline to identify VCFs in a straightforward and intuitive way.  Our model generates rules based on domain knowledge about vertebral anatomy and fracture characteristics, ensuring that the decision-making process is interpretable and evidence-based.\cite{guideline2022, Genant2003, Lenchik2004AJR}

\begin{figure}[tbp]
    \centering
    \subfloat[Height Map Extraction]{
        \includegraphics[width=0.75\linewidth]{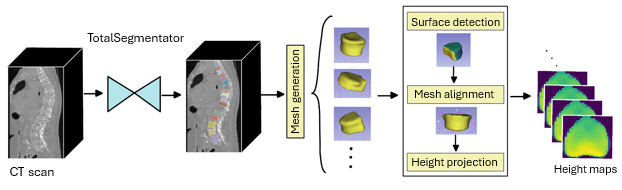}
        \label{fig:overview}
    }
    \hfill
    \subfloat[Height Map Sections]{
        \includegraphics[width=0.19\linewidth]{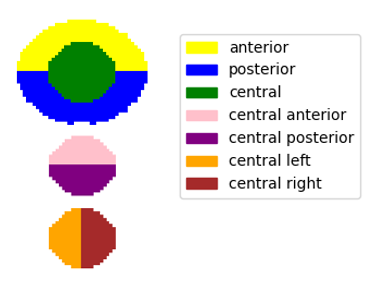}
        \label{fig:sections}
    }
    \caption{\protect\subref{fig:overview} The height map extraction process relies on vertebral body segmentation with TotalSegmentator~\cite{wasserthal2023totalsegmentator} to obtain height maps in a consistent local coordinate system. \protect\subref{fig:sections} Drawing on real diagnostic decision-making, we define 7 regions of the vertebral body to inform the severity rating.}
    \label{fig:heightmap}
\end{figure}

Figure \ref{fig:overview} shows the pipeline followed to generate the vertebral shape measurements. First, for every CT scan, the vertebral bodies are segmented using TotalSegmentator~\cite{wasserthal2023totalsegmentator, isensee2021nnu}. Then, individual 3D meshes are generated for each vertebral body using the marching cubes algorithm\cite{LorensenMarchingCubesAlgo}. Height map generation consists of multiple steps: main surface detection by applying k-mean clustering (k=6) on the point cloud normal vectors; mesh reorientation to ensure the posterior and the inferior surfaces are aligned with their corresponding planes; height projection by fitting a 3D grid cell to the point cloud and computing the maximum height per column. Finally, seven regions of interest (ROIs) are defined and their statistical measurements (mean and standard deviation) are extracted. Similarly to traditional approaches that utilize the Anterior-Posterior ratio (APR), Middle-Posterior ratio (MPR), and Middle-Anterior ratio (MAR) to quantify bone height loss\cite{Baum2014VCF}, we compute the pair-wise ratio of our section's average heights. Conventionally, radiologists compare the vertebral heights (AH, MH, and PH) with an anatomically proximate comparator to determine the percentage height loss \cite{RSNA2017VCF}. For each vertebra, we quantified this observation by finding the vertebra with the highest central average height (\textbf{C}) and computing the section-wise ratios between both height maps (\textbf{$\overline{C}$}).

\subsection{Rule generation}
\label{sec:RuleGen}

Our model uses the RuleFit\cite{rulefit} algorithm to identify the optimal combination of feature thresholds that enhance accuracy. RuleFit combines decision rules from tree-based models with linear regression to create a predictive model that can capture non-linear relationships. Given a tree ensemble, Rulefit creates rules from all the trees of the ensemble, with each rule defining specific conditions based on feature thresholds (Eq \ref{eq:rule}).
\begin{equation}
r_m(x) = \prod_{j\epsilon T_m}I(x_j \epsilon s_{jm}),
\label{eq:rule}
\end{equation}
where \(T_m\) is the set of features used in the m-th tree, I is the indicator function that is 1 when feature \(x_j\) is in the subset of values s for the j-th feature and 0 otherwise.
In our case, the tree ensemble model used is GradientBoostingClassifier, and the 3 rules with the highest stratification power are included as features for the regression model. Finally, we train a sparse linear model with LASSO on the new rule features, which results in a linear parametrization of the model,
\begin{equation}
\hat{f}(x) = \hat{\beta}_0 + \sum^K_{k=1}{\hat{\alpha _k}r_k(x)},
\label{eq:linear}
\end{equation}
where \(\hat{\alpha}\) is the estimated weight vector for the rule features and \(\hat{\beta_0}\) is the intercept.

\section{RESULTS}
\label{sec:results}

\paragraph{Dataset} 
Our data was obtained from the Large Scale Vertebrae Segmentation Challenge (VerSe19)\cite{verse19} challenge and the ground truth scores defined by the Genant semiquantitative grading system\cite{verseAnnot} are publicly available as well. The VerSe19 dataset originally included 160 CT scans from 141 patients, containing 1491 vertebrae. In this study, vertebrae with foreign material such as screws and other metal prostheses were excluded. Moreover, scans with only one vertebrae annotation were not used, as the absence of another vertebra impeded the calculation of intervertebral ratios. After these exclusions, the final dataset contains 1,460 annotated thoracolumbar vertebrae. The dataset followed the same train/validation/test split defined in \cite{sekuboyina2020verse}.

\paragraph{Rules for VCF detection}
RuleFit returns a set of coefficients associated with each binary rule. According to the coefficients provided by our trained linear model, the VCF prediction is determined by Eq.~\ref{eq:linear}, resulting in the dot product,
\begin{equation}
\hat{f}(x) = 0+ \begin{bmatrix}
1.49471001 & 0.36870275 & -4.10884354
\end{bmatrix}
\cdot
\begin{bmatrix}
\mathtt{avg(A_0):avg(P)} \leq 0.91 \\
\mathtt{avg(A_0):avg(P)} > 0.91 \& \mathtt{avg(C):avg(\overline{C})} \leq 0.81 \\
\mathtt{avg(A_0):avg(P)} > 0.91 \& \mathtt{avg(C):avg(\overline{C})} > 0.81
\end{bmatrix}.
\label{eq:linear_model}
\end{equation}
Here, \textbf{A$_{0}$} is antero-centric section, \textbf{P} is the posterior section, \textbf{C} is the whole central section, \textbf{$\overline{C}$} is the central section of the reference vertebra mentioned in Section \ref{sec:overview}, and \textbf{avg(X)} indicates the average height of section X. Thus, a VFC will be predicted positive \textbf{IF}: the average height of the antero-centric section is smaller than 91\% of the average height of the posterior section \textbf{OR} that condition is not met but the average height of the central section is smaller than the 81\% of the average height of the reference-vertebra central section. 
\\
\paragraph{Black-box model benchmark}
We trained two backbone deep learning models, ResNeXt50 and DenseNet, as a benchmark in our study. For each vertebra, we resampled a cropped volume with 1mm isotropic resolution. A stack of 14 centered sagittal slices was then extracted. Augmentation consisted of random rotations and flips. A class-balanced data sampler was used during training, based on all four fracture levels. After training, the slice-level predictions and labels are binarized. The slice-level threshold that determines the vertebra-level predictions is obtained by optimizing the Youden J statistic index during validation.\cite{Youden}. Similar to \cite{verseVCFDet}, we take advantage of the available vertebral body segmentations to mask the 2D samples and retrain the DL models.

\begin{figure}[tbp]
    \centering
    \subfloat[Neurosymbolic Decision Process]{
        \label{fig:visComp}
        \includegraphics[height=1.8in]{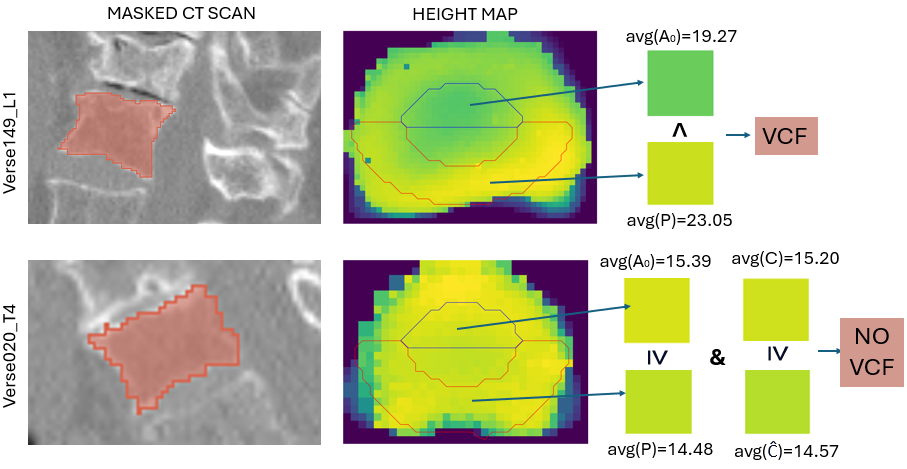}
    }
    \hfill
    \subfloat[Example Cases]{
        \label{fig:CM}
        \includegraphics[height=1.35in]{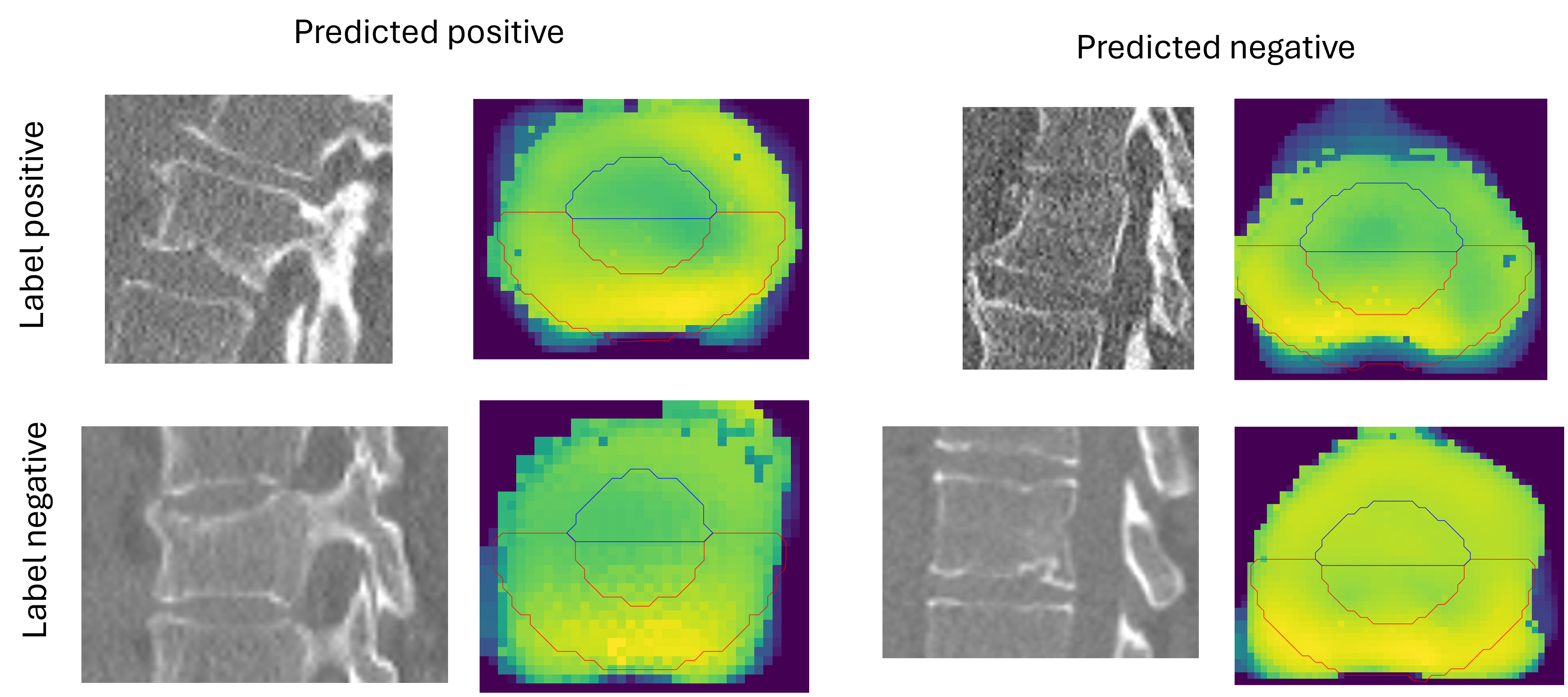}
    }
    \caption{\protect\subref{fig:visComp} Our neurosymbolic reasoning strategy evaluates the relative height of the vertebral body in specific regions to identify cases of VCF. \protect\subref{fig:CM} Example model output. A common failure mode occurs when the vertebral body's shape suggests a compression fracture, even if one is not labeled.
    }
\end{figure}

We evaluate every model on VerSe19 test set (Table \ref{tab:performance}). Both DL models benefit from masked data which can be explained given the importance of detecting the vertebral endplates in the prediction of VCFs. However, our neurosymbolic approach outperforms both ResNeXT and DenseNet with and without masking.
Figure \ref{fig:visComp} and \ref{fig:CM} highlight the interpretability of the proposed model by providing a clear and intuitive visual representation of its reasoning criteria. This contrasts with the more complex raw sagittal CT images and vertebra masks. The generated maps facilitate visual inspection of data during opportunistic screenings by simplifying the assessment process, enabling quicker and more straightforward evaluations of vertebral conditions

\begin{table}[t]
    \centering
    \small %
    \setlength{\tabcolsep}{4pt} %
    \begin{tabular}{lcccc}
        \toprule
        \textbf{Model} & \textbf{F1} & \textbf{Accuracy} & \textbf{Precision} & \textbf{Recall} \\
        \midrule
        ResNeXt\_unmasked & 0.62 & 0.88 & 0.45 & 0.97 \\
        ResNeXt\_masked & 0.65 & 0.9 & 0.5 & 0.94 \\
        DenseNet\_unmasked & 0.65 & 0.91 & 0.5 & 0.94 \\
        DenseNet\_masked & 0.77 & 0.95 & 0.67 & 0.91 \\
        Interpretable approach & \textbf{0.81} & \textbf{0.96} & \textbf{0.74} & \textbf{0.91} \\
        \bottomrule
    \end{tabular}
    \caption{Vertebra-level performance metrics comparison}
    \label{tab:performance}
\end{table}

\section{CONCLUSION}
\label{sec:conclusion} 
Our neurosymbolic rule-based model demonstrates superior performance in detecting VCFs in CT scans compared to traditional deep learning models. Its intrinsic transparency provides clear visual and logical cues for the decision-making process, which is especially appealing in the scope of opportunistic detection of VCFs. Future work will investigate whether our explainable approach will allow users to adequately calibrate their trust in the automated detection method, thus supporting usability. 

\printbibliography

\end{document}